\documentclass[sigconf]{acmart}
\usepackage{ulem}  % for strikethrough

\newtheorem{dfn}{Definition}
\usepackage{tabularx}  % for full page table (?)
\usepackage{algorithm, algpseudocode}
\usepackage{enumitem}
\usepackage{subcaption}
\usepackage{mwe}
\usepackage{wrapfig}
\usepackage[colorinlistoftodos]{todonotes}
\usepackage{multirow}
\usepackage{amsmath,amsthm}
\usepackage{mathrsfs, bbm, tikz-cd}
\usepackage{bm}
\usepackage{float}
\usepackage{mathtools}

\theoremstyle{plain}

\newcommand{\cB}{\mathcal{B}}

\newcommand{\cL}{\mathcal{L}}
\newcommand{\cP}{\mathcal{P}}

\newcommand{\cN}{\mathcal{N}}

\newcommand{\cX}{\mathcal{X}}
\newcommand{\cY}{\mathcal{Y}}

\newcommand{\cR}{\mathcal{R}}

\newcommand{\R}{\mathbb{R}}

\DeclareMathOperator*{\argmin}{\mathrm{arg\,min}}

\newcolumntype{Y}{>{\centering\arraybackslash}X}
\newlist{todolist}{itemize}{2}
\setlist[todolist]{label=$\square$}

\AtBeginDocument{%
  \providecommand\BibTeX{{%
    \normalfont B\kern-0.5em{\scshape i\kern-0.25em b}\kern-0.8em\TeX}}}

\setcopyright{acmcopyright}
\copyrightyear{2022}
\acmYear{2022}
\acmDOI{XXXXXXX.XXXXXXX}

\acmConference[KDD '22]{Make sure to enter the correct
  conference title from your rights confirmation emai}{August 14--18, 2022}{Washington, DC}
\acmPrice{15.00}
\acmISBN{978-1-4503-XXXX-X/18/06}

\begin{document}

%%
%% The "title" command has an optional parameter,
%% allowing the author to define a "short title" to be used in page headers.
\title{BASED-XAI: Breaking Ablation Studies Down for Explainable Artificial Intelligence}

%%
%% The "author" command and its associated commands are used to define
%% the authors and their affiliations.
%% Of note is the shared affiliation of the first two authors, and the
%% "authornote" and "authornotemark" commands
%% used to denote shared contribution to the research.
\author{Isha Hameed}
\affiliation{%
  \institution{Capital One - C4ML}
%   \streetaddress{P.O. Box 1212}
  \city{Boston}
  \state{MA}
  \country{USA}
%   \postcode{43017-6221}
}
\email{isha.hameed@capitalone.com}

\author{Samuel Sharpe}
% \authornote{Both authors contributed equally to this research.}
% \orcid{1234-5678-9012}
\affiliation{%
  \institution{Capital One - C4ML}
%   \streetaddress{P.O. Box 1212}
  \city{Boston}
  \state{MA}
  \country{USA}
%   \postcode{43017-6221}
}
\email{samuel.sharpe@capitalone.com}

\author{Daniel Barcklow}
\affiliation{%
  \institution{Capital One - C4ML}
%   \streetaddress{P.O. Box 1212}
  \city{Washington}
  \state{DC}
  \country{USA}
%   \postcode{43017-6221}
}
\email{daniel.barcklow@capitalone.com}

\author{Justin Au-Yeung}
\affiliation{%
  \institution{Capital One - C4ML}
%   \streetaddress{P.O. Box 1212}
  \city{Boston}
  \state{MA}
  \country{USA}
%   \postcode{43017-6221}
}
\email{justin.au-yeung@capitalone.com}

\author{Sahil Verma}
% \authornotemark[1]
\affiliation{%
  \institution{Capital One - C4ML}
%   \streetaddress{P.O. Box 1212}
  \city{San Francisco}
  \state{CA}
  \country{USA}
%   \postcode{43017-6221}
}
\email{sahil.verma@capitalone.com}

\author{Jocelyn Huang}
% \authornotemark[1]
\affiliation{%
  \institution{Capital One - C4ML}
  \streetaddress{P.O. Box 1212}
  \city{New York}
  \state{NY}
  \country{USA}
  \postcode{43017-6221}
}
\email{jocelyn.huang@capitalone.com}

\author{Brian Barr}
% \authornotemark[1]
\affiliation{%
  \institution{Capital One - C4ML}
  \streetaddress{P.O. Box 1212}
  \city{New York}
  \state{NY}
  \country{USA}
  \postcode{43017-6221}
}
\email{brian.barr@capitalone.com}

\author{C. Bayan Bruss}
% \authornotemark[1]
\affiliation{%
  \institution{Capital One - C4ML}
%   \streetaddress{P.O. Box 1212}
  \city{Washington}
  \state{DC}
  \country{USA}
%   \postcode{43017-6221}
}
\email{bayan.bruss@capitalone.com}

%%
%% By default, the full list of authors will be used in the page
%% headers. Often, this list is too long, and will overlap
%% other information printed in the page headers. This command allows
%% the author to define a more concise list
%% of authors' names for this purpose.
\renewcommand{\shortauthors}{Hameed, et al.}

%%
%% The abstract is a short summary of the work to be presented in the
%% article.
\begin{abstract}
  Explainable artificial intelligence (XAI) methods lack ground truth.  In its place, method developers have relied on axioms to determine desirable properties for their explanations' behavior.  For high stakes uses of machine learning that require explainability, it is not sufficient to rely on axioms as the implementation, or its usage, can fail to live up to the ideal. As a result, there exists active research on validating the performance of XAI methods. The need for validation is especially magnified in domains with a reliance on XAI. A procedure frequently used to assess their utility, and to some extent their fidelity, is an \textit{ablation study}. By perturbing the input variables in rank order of importance, the goal is to assess the sensitivity of the model's performance. Perturbing important variables should correlate with larger decreases in measures of model capability than perturbing less important features. While the intent is clear, the actual implementation details have not been studied rigorously for tabular data. Using five datasets, three XAI methods, four baselines, and three perturbations, we aim to show 1) how varying perturbations and adding simple guardrails can help to avoid potentially flawed conclusions, 2) how treatment of categorical variables is an important consideration in both post-hoc explainability and ablation studies, and 3) how to identify useful baselines for XAI methods and viable perturbations for ablation studies.
\end{abstract}
% \linepenalty=100
%%
%% The code below is generated by the tool at http://dl.acm.org/ccs.cfm.
%% Please copy and paste the code instead of the example below.
%%


%%
%% Keywords. The author(s) should pick words that accurately describe
%% the work being presented. Separate the keywords with commas.

%% A "teaser" image appears between the author and affiliation
%% information and the body of the document, and typically spans the
%% page.
\begin{teaserfigure}
  \includegraphics[width=\textwidth]{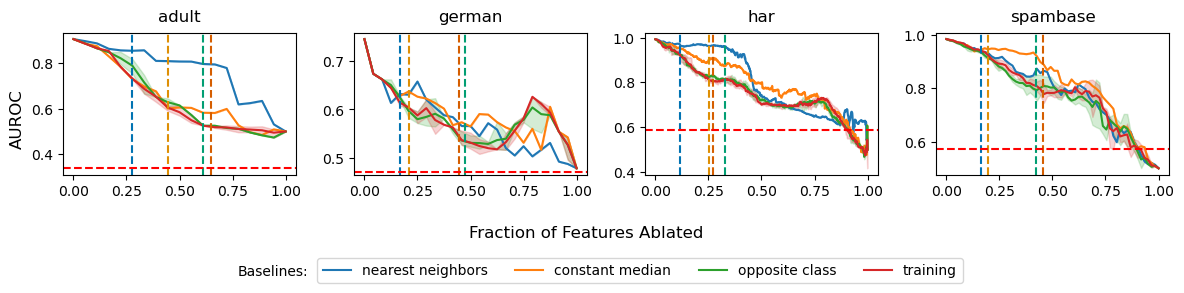}
  \caption{Comparison of baselines with a constant median perturbation}
  \Description{This may be the best result of the paper.}
  \label{fig:teaser}
\end{teaserfigure}

% \todo[inline]{added aspect ratio to plot (teaser_v2.png) - looked kind of small, this is with no aspect ratio changes }

%%
%% This command processes the author and affiliation and title
%% information and builds the first part of the formatted document.
\maketitle

\section{Introduction}
Along with the spectacular performance of black box models on complex data sets, comes a need to explain how those models make their predictions, whether the motivation for explanations originate with an internal responsibility from the model developers, or is encouraged by policy~\cite{eu_guidelines}. Explaining predictions of black box models typically comes from the perspective of local attributions, providing some measure of importance for each feature of all data observations of interest.

% black box models on sufficiently complex tabular data
A lack of comprehensive ground truth sources for local attributions leads to an inadequate understanding of which methods and hyperparameters are best for a particular use case (e.g. global feature selection,  XAI baseline selection, and local explanations). \textit{Ablation studies} attempt to assess the effectiveness of a set of local and global attributions through sensitivity of model capability under a given perturbed input. Perturbing inputs deemed important by an XAI method should lead to larger decreases in model performance than perturbing less important features. This intuitive behavior can be further expanded to the shape of the ablation curve: the drop in model capability as a function of the fraction of ablated input features. The value of the curve, as well as its derivative, should be monotonically decreasing as the fraction of ablated features increase.

We focus on applying ablation to models, built on a variety of open source tabular datasets with varying feature types, to assess the capability of XAI methods for tabular data. To broaden the scope of XAI methods available for experimentation, we use differentiable models, although the process in general translates to non-differentiable models (e.g. gradient boosted trees).

Our contributions are as follows:
\begin{itemize}
    \item We differentiate ablation perturbations vs XAI baseline distributions by examining the effect of changing both in our experiments. This removes the confounding effect of using the same perturbation across experiments and matching baseline and perturbation distributions.
    % \item We identify opportunities during the processes of interpreting the explanations and perturbing features for ablation where categorical features must be carefully treated.
    \item We identify the need to treat categorical features in their label encoded form when applying perturbations \textemdash as opposed to the one hot encoded form used as model inputs. This treatment provides a more accurate ranking of features and helps keep ablation perturbations in-sample.
    \item We propose \textit{guardrails} or sanity checks for ablation studies that help to define the feasible region of an ablation study. This guides the attention away from problematic regions and results in firmer conclusions.
    
\end{itemize}

Taken together, these contributions provide a more rigorous approach for conducting ablation studies on tabular data and raise further questions for future work.
\section{Previous work}
%fighting with LaTeX on figure placment - this really belongs with 
%Section 3 Method
\begin{figure*}[!ht]
  \includegraphics[width=\textwidth]{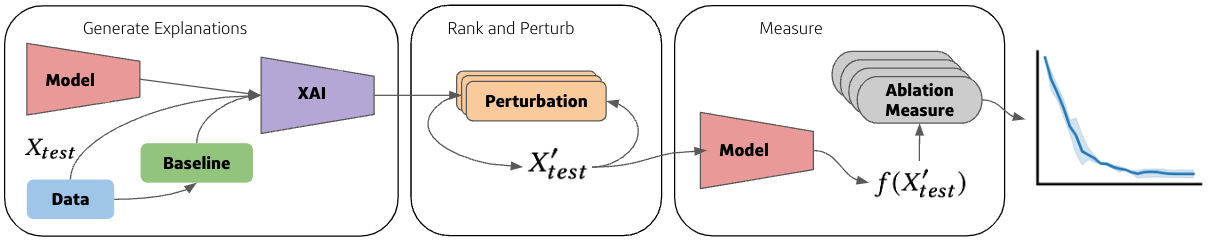}
  \caption{Diagram of data and logic flow for an ablation study. The process starts with a trained model, an XAI method, and a reference distribution generated from training data called a baseline (see Table~\ref{tab:perturb_baseline}). Features of test samples to be explained are rank ordered and perturbed based on importance and  assessed with model capability. The series of ablation steps are summarized in an ablation curve (far right).}
  \label{fig:diagram}
\end{figure*}

\subsection{Post-hoc explainability}

\textit{Post-hoc explainability} is part of a diverse, collective effort to explain high performing black box models. Global, local, model-agnostic, and model-specific methods each attempt to address post-hoc explainability, but a common framework for application and validation of these methods has not yet been established.

Gradient-based attribution is one model-specific sub-field within post-hoc explainability that seeks to generate attributions for neural network inputs.  Common to other XAI methods, it is difficult to determine if, and under which conditions, these gradient-based explanations are valid. \citet{adebayo2018} demonstrate through randomization tests that there are certain conditions where popular gradient-based methods appear independent to training data permutations and model parameter changes. They show that the combination of these inadequate methods with misleading visualizations underscore the need for guidelines and sanity checks \cite{adebayo2018}. Guidance is not only needed for gradient-based attribution methods, but also for popular XAI methods such as SHAP.

SHAP is a local additive feature attribution method developed to align with classical cooperative game theory methods. \citet{lundberg2017unified} propose several implementations of SHAP: Kernel SHAP and Deep SHAP. Kernel SHAP was designed to be an efficient model-agnostic estimation of Shapley values while the model-specific Deep SHAP leverages work from DeepLIFT to more accurately estimate Shapley values for neural network models \cite{lundberg2017unified}.  \citet{slackfoollimeshap2020} present an adversarial method for developing a biased classifier that cannot be detected by SHAP, highlighting vulnerabilities and the need for validation frameworks within various sub-fields of XAI. The lack of consensus on an XAI validation framework results in misapplication of XAI methods and misinterpretation of results.  

There are efforts to move away from traditional XAI methods that naively permute or vary features without considering interactions between them. \citet{hooker2021extrapolation} categorize these as "permute-and-predict (PaP)" methods. They show that feature correlation biases attributions and suggest augmenting these PaP approaches by either
sampling features conditionally\footnote{This method is often applied to tree based models where conditional sampling is feasible.}, thereby remaining within the model's in-sample operating range, or by dropping features and retraining. Similar fundamental criticisms are made of ablation studies.

\subsection{Use of ablation to assess XAI methods}

Ablation studies assess XAI methods by ablating or removing feature information from observations based on importance ranking. If the XAI method is appropriately applied, features deemed important will cause model capability to decrease when ablated. \citet{ROAR} were early adopters of using ablation to interrogate explanations for image processing models. Working with ImageNet data and RESNET-50 models, they ablated features sequentially by replacing a fraction of the important pixels with the mean value and retraining the model.  During this process, \citet{ROAR} measured model accuracy after being retrained with ablated features to assess the XAI method employed. Also within the field of XAI for image processing, \citet{sturmfels2020visualizing} investigated the impact of a broader set of baselines on explanations by measuring changes in an ImageNet classifier's prediction confidence when ablating pixels based on attributed importance. Outside of the field of image processing, there is research to better understand models that process tabular data.

% ? Not sure what to do with this.  It was here before adding text. - Dan B.
%
% Their method enables measurement of an XAI method or ensemble's capability to explain above that of a randomized local explanation. 

% Adding to the body of work using ablation for image classification, \citet{sturmfels2020visualizing} investigates the impact of a broader set of baselines on explanations for ImageNet classifier predictions. 

With an emphasis on tabular data, \citet{haug2021baselines} introduce a taxonomy of baseline methods for attribution models to aid the selection of baselines in research and practice.  \citet{haug2021baselines} discovered through ablation studies that baselines have a significant impact on generated feature attributions. 
Furthermore, they identified that, although there were baselines that consistently failed to produce valuable feature attributions, there were no universally superior baselines \cite{haug2021baselines}. They hypothesize that baseline performance may be dependent on the baseline's ability to approximate the original data generating distribution.  Specifically, feature attributions may be more discriminative the closer a baseline approximates the original data generating distribution \cite{haug2021baselines}.  This finding suggests that baselines that deviate out-of-distribution (OOD) produce invalid explanations that are reflected in ablation curves. Therefore, sanity checks for ablation studies could be developed to guide baseline selection for an attribution method using tabular data.

\subsection{Weaknesses and criticisms of ablation}

Previous work discusses and attempts to address problems with OOD issues typically present in ablation studies and feature removal based explanation methods in general \cite{ROAR, covert2021explaining,hase2021out}.  \citet{ROAR} state that "without retraining it is unclear whether the degradation in model performance comes from the distribution shift or because the features that were removed are truly informative". However, we have observed a lack in consensus on the appropriate way to interpret fine-grained explanations yielded from retrained surrogate models. Although retraining-based methodologies diverge from the post-hoc paradigm, there are some fair criticisms to be made of ablation studies.  

While the work of \citet{haug2021baselines} extends ablation to a more thorough investigation on tabular data, we believe that there are areas where their experiments lack thoroughness. We believe that their baseline methods that were derived from the image processing domain are inappropriate for tabular data and result in OOD data. Additionally, they use a small sample size of 10 for sample-based baselines, which is not supported and fairly small relative to common practice. Most importantly, they only use one perturbation distribution (Gaussian blur) for all experiments. Not only does this conflate the effect of baseline choice in XAI methods, but Gaussian blur can cause extreme OOD data in the tabular domain.

\section{Method}

% \begin{figure*}[h!]
%   \includegraphics[width=\textwidth]{figures/section_3/diagram_box.png}
%   \caption{Diagram of data and logic flow for an ablation study. The process starts with a trained model, an XAI method with baseline, and a set of samples to be explained, progresses to rank ordered perturbations based on importances, ending with an assessment of model capability.  The series of ablation steps are summarized in an ablation curve (far right).}
%   \label{fig:diagram}
% \end{figure*}

\subsection{Overview}
In this section we describe the process of an ablation study given a trained classification model $f:\cX \to \cY$, with $\cX = \R^{n \times d}$ and $\cY = \R^{n \times c}$, and $n$ samples, $d$ features, and $c$ classes.  We refer to single inputs $x_i \in \cX \; \forall i \in \{1\dots n\}$ and their features as $x^j \; \forall j \in \{1\dots d\}$.  

An ablation study relies on explanations originating from an XAI method and its baseline as well as perturbations. 

\begin{dfn}
    An explanation method is any local feature attribution or global importance method $A$ that attributes the output of $f$ for a specific observation $x_i$ or assigns importance to each feature for a collection of observations. 
\end{dfn}

There are a wide variety of explanation methods that exist. We use methods in our evaluation which require baseline distributions. 

\begin{dfn}
    Baselines are reference distributions used in general removal or gradient based explanation methods. Baselines can be default values or a distribution over which removed features are marginalized. 
\end{dfn}

\begin{dfn}
    A perturbation $p \sim \cP$ attempts to remove information from $x_i^j$ by replacing $x_i^j$ with $p_i^j$. 
\end{dfn}

\begin{dfn}
    An ablation study attempts to assess the validity of local or global explanations through examining the relative model performance change as input features $x^j$ are perturbed in order of importance. 
\end{dfn}

\subsection{Models and explanation methods}
We train differentiable models for each ablation study to expand our set of available explanation methods. Each model is a two layer neural network with hidden neurons proportional to the number of features in each dataset. We select the best performing model with early stopping on held out validation data. 

% \subsection{Explanation Methods}

% We use three different explanation methods in our ablation studies. These methods include model agnostic, gradient based, and Shapley value based methods. 

% \begin{itemize}
%     \item Deep SHAP (DS) \cite{lundberg2017unified}
%     \item Integrated Gradients (IG) \cite{SundararajanIG2017}
%     \item Kernel SHAP (KS) \cite{lundberg2017unified}
% \end{itemize}

We use two gradient based explanation methods, Deep SHAP~\cite{lundberg2017unified} and Integrated Gradients~ \cite{SundararajanIG2017}, along with one model agnostic method, Kernel SHAP\cite{lundberg2017unified}. We use these methods to generate local explanations across all datasets and baselines. Global explanations are taken to be the average absolute value across the local explanations.

\subsection{Baseline and perturbation distributions}
Our studies use the following set of sampling methods, listed in Table~\ref{tab:perturb_baseline}, as either baseline samples or perturbations during ablation.

\begin{table}[h] 
    \begin{tabular}{cccc} 
     \toprule[1pt]
    Distribution      & Perturbation      & Baseline   & Reference \\
    % \midrule
    \hline
    constant median        & \checkmark    & \checkmark  &   \\ 
    marginal distribution  & \checkmark    &             & \cite{breiman2001}                            \\
    nearest neighbors &               & \checkmark  &  \cite{albini2021counterfactual}\\              
    opposite class    &               & \checkmark  &    \cite{albini2021counterfactual}\\      
    training          &               & \checkmark  &  \cite{lundberg2017unified,haug2021baselines}   \\
    max distance      & \checkmark    &   & \cite{haug2021baselines,sturmfels2020visualizing} \\
    \hline
    \end{tabular}
    \vspace{0.1cm}
\caption{\label{tab:perturb_baseline} Perturbations and baselines used in our experiments.}
\vspace{-0.6cm}
\end{table}

The \textit{constant median} distribution is a variant of the constant distribution which is simply the median of each feature. We believe this is a more meaningful baseline than a constant zero baseline for tabular data. 

We use the \textit{marginal} distribution as a perturbation replacing each perturbed feature with a value drawn from the training marginal distribution of that feature~\cite{breiman2001}.

The \textit{training} baseline,  also known as the expectation baseline, is a sub-sample of the training data \cite{lundberg2017unified}. 

We borrow two baselines from \citet{albini2021counterfactual} that introduce alternative counterfactual baselines\textemdash opposite class and nearest neighbors. The \textit{opposite class} baseline includes a sample from the training data with an observed class that is opposite of the predicted class of the explained observation. The \textit{nearest neighbor} baseline selects $k$ samples that are closest to the data sample. For this study the value of $k$ is fixed at five. Finally, we use the \textit{max distance} perturbation \cite{sturmfels2020visualizing} as a strident contrast to the the median and marginal perturbations.

\subsection{Hyperparameters for experiments}

\begin{figure}[!ht]
  \includegraphics[width=\columnwidth]{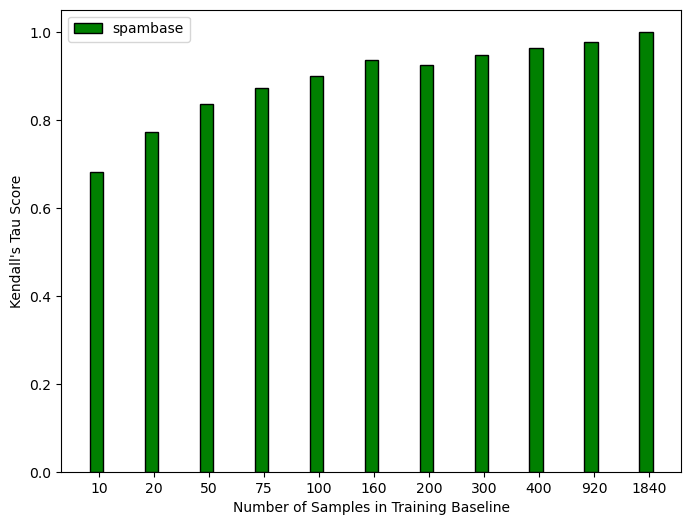}
  \caption{To determine a sufficient baseline sample size we compare Kendall Tau correlation of global explanation rankings at varying baseline sample sizes relative to the full training dataset (1840 samples) using Deep SHAP explanations.
  }
  \label{fig:kt_nsample_selection}
\end{figure}

To select a common sample size for sampling baselines (training, opposite class) we examine how the rankings of global explanations change as we vary the size of the baseline sample. We measure the change in rankings by using Kendall's Tau to compare each set of rankings with the rankings derived from explanations using the full training data as a baseline. Over all datasets we find that a sample size of 50 captures sufficiently similar rankings to the full training data. The analysis for Spambase can be seen in Figure~\ref{fig:kt_nsample_selection}.

We perform a similar analysis for the larger datasets to ensure that our experiments are manageable with reduced compute resources yet still robust. We use a 50\% stratified subsample of the Human Activity Recognition (har) and adult datasets in our experiments.

%   Variation of the Kendall Tau correlation coefficient for feature ranking across different sample sizes for the training baseline.

\subsection{Ablation study algorithm}

We detail the steps of an ablation study based on local explanations in Algorithm~\ref{alg:ablation}. Ideally, ablation studies expect that any good explanation method will return explanation values that adhere to the following property: features removed from the model with higher importances decrease model performance more than those with lower importances. 

\begin{algorithm}
\caption{Ablation study}
\label{alg:ablation}
\begin{algorithmic}[1]
\State Given a model $f$, evaluation data $X, Y$, perturbation distribution $\cP$, baseline distribution $\cB$, explanation method $A$, and an ablation performance metric $\cL$, and number of trials $T$.
\State Initialize $scores$ to track loss metric
\For{t $\leftarrow  1 \dots T$}
\State Sample $b \sim \cB$ and $p \sim \cP$.
\For{$i \leftarrow \{1 \dots n\}$}
\State Compute explanations $e_i$ via method $A$ and baseline $b$ 
\State Compute ranking $r_i \leftarrow argsort(e_i)$ in decreasing order
\State where $r_i^k$ is index of the $k$th ranked feature.
\EndFor
\State $scores(t, 0) \leftarrow  \cL(f(X),Y)$ \Comment{Loss on original $X$}
\For{$k \leftarrow \{1 \dots d\}$} 
\For{$i \leftarrow \{1 \dots n\}$}
\State $m \leftarrow r_i^k$  \Comment{$k$th ranked feature for $x_i$}
\State $x_i^{m} \leftarrow p_i^{m}$ \Comment{perturb $x_i$}
\EndFor
\State $scores(t, k) \leftarrow \cL(f(X),Y)$
\EndFor
\EndFor
\State \Return scores
\end{algorithmic}
\end{algorithm}

\subsection{Sanity checks}
\label{sanity_checks}
% full page width version

\begin{figure}[h]
  \begin{subfigure}{.5\columnwidth}
    \centering
    \includegraphics[width=\columnwidth]{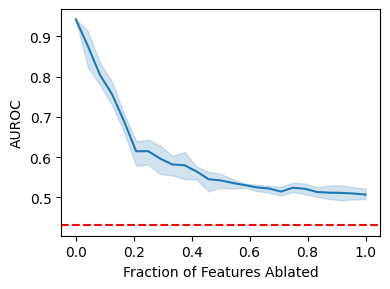}
    \caption{Worst case model}
    \label{fig:sanity_worst_model}
  \end{subfigure}%
  \begin{subfigure}{.5\columnwidth}
    \centering
    \includegraphics[width=\columnwidth]{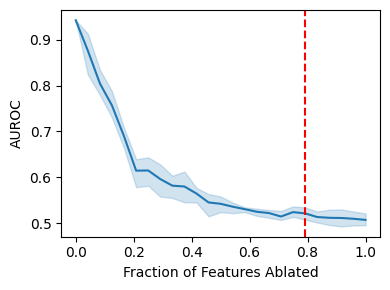}
    \caption{Random features}
    \label{fig:sanity_random_feature}
\end{subfigure}
  
\begin{subfigure}{.49\columnwidth}
    \centering
    \includegraphics[width=\columnwidth]{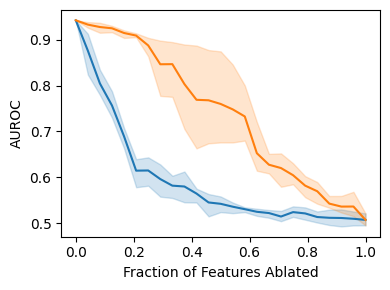}
    \caption{Random explanations}
    \label{fig:sanity_random_exp}
\end{subfigure}
\begin{subfigure}{.49\columnwidth}
    \centering
    \includegraphics[width=\columnwidth]{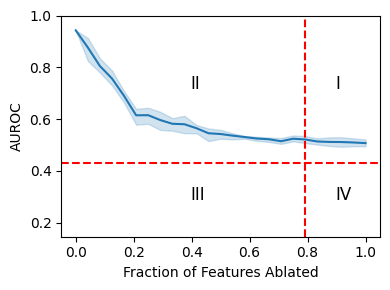}
    \caption{Guardrail quadrants}
    \label{fig:sanity_quadrants}
\end{subfigure}
  \caption{Examples of the three guardrails (a, b, c) and their quadrants (d) overlaid on an ablation experiment.}
\end{figure}

In addition to the ablation procedure outlined above, three sanity checks are overlaid on the plots and are used as guidance in understanding the results. The first sanity check is a model trained with shuffled labels to provide a lower bound on model capability. We suspect that any model performance that drops below this control may have degraded due to other factors such as out of distribution inputs. We display this guardrail as a horizontal dashed line as seen in Figure~\ref{fig:sanity_worst_model}.

% \begin{figure}[!ht]
%   \includegraphics[width=0.5\columnwidth]{figures/section_3/sanity_check_worst_case_model.png}
%   \caption{Illustration of the worst case model sanity check (horizontal dashed line).}
%   \label{fig:sanity_worst_model}
% \end{figure}

The second sanity check attempts to put a lower bound on the value of a feature's explanation. We augment the original dataset with four additional randomly generated features  $x_i \sim \cN(0,1)$. These random features are not correlated with the prediction target, thus should have no importance. Features with attributions smaller in absolute value than the added random features are likely not important, and we should not focus on changes in model performance after perturbing these features. For global explanations, we assign this sanity check as the rank of the most important random feature. When using local explanations, we take the best average rank of the random features. Given an ordinal rank $o_i^j$ for sample $x_i$ at feature $j$ and a set of random features $\cR$, we calculate the guardrail as:
$$\argmin_{j \in \cR} \frac{1}{N}\sum_{i=1}^N o_i^j $$

To represent this sanity check, we add a dashed vertical line as seen in Figure~\ref{fig:sanity_random_feature}.

Similar to \citet{haug2021baselines}, we also use a random set of explanations, ablating features in a uniformly random order, as a reference to benchmark against. As shown in Figure~\ref{fig:sanity_random_exp}, the random ordering results in a steady decline in model capability. 

% \subsection{MISC SCRATCH SPACE}
% Questions raised by our definition the we need to be able to answer:
% \begin{enumerate}
%     \item How do we evaluate if we have minimized a feature's importance? Or phrased another way - what makes a good ablation method?
%     \begin{itemize}
%         \item Is rerunning the predictions on the dataset with the ablated feature sufficient?  
%         \item If we normalized the global importances so they summed to one, is the best ablation method the one that drops model performance in proportion to the feature importance?
%         \item What if we find a method that drops model performance further than expected?  Is that a better ablation method or a bad set of explanations?
%     \end{itemize}   
    
%     % \item Should it be just one feature at a time?  In the presence of correlated variables - would ablating principal components be "better"?
%     % \item \sout{Should we retrain the model having dropped it entirely?  (this would be an improvement on ROAR...which ablated pixels, but never tested their assumption that the ablated pixels had no importance)} - a good question to answer, but probably not in a workshop paper - will add in for a conference paper
% \end{enumerate}

\subsection{Categorical treatment}
\label{section:cat}
Previous work often does not address tabular data, and when it does, categorical variables do not have their own distributional assumptions in baselines or perturbations~\cite{haug2021baselines}. We propose two alterations to the ablation process in the presence of categorical variables. 
\begin{enumerate}
    \item For ranking, each categorical feature has a single ranking derived from the sum of attributions of its one-hot-encoded features. 
    \item For perturbations, categorical features are carefully perturbed such that the result is a valid category from the original training distribution.  For example, constant median perturbation for a categorical feature is modified to the most common category.
\end{enumerate}

\begin{figure}[ht]
  \includegraphics[width=\columnwidth]{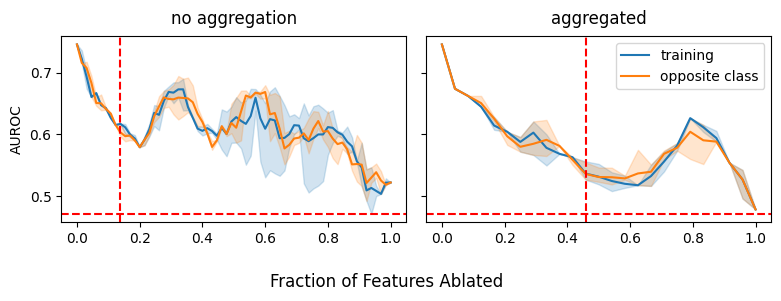}
  \caption{ An example of how our categorical treatment leads to smooth ablation curves with less noise caused by one-hot encoded perturbations. On the left we leave categorical variables in one-hot form when perturbing and ranking while on the right we transform them to their categorical form. }
  \label{fig:german_agg_comparison}
\end{figure}

By combining the one-hot-encoded explanations, we observed that our vertical guardrail sanity check moves to the right. This can be observed in Figure~\ref{fig:german_agg_comparison}.

%  As a consequence, we alter constant median perturbation for categorical variables as replacement with the most common category. Additionally, since 

% Performing perturbations in the multinomial category space is feasible for only the \textit{marginal} distribution, so we alter definitions for the other distributions.

Since performing perturbations in the multinomial category space is feasible for only the marginal distribution, we additionally alter max distance. Max distance in continuous space is the furthest point in $L_1$ space, so we sample uniformly from the other categories. Formally, for a categorical feature $x_i^j = a$ with categories $C$ as: 
$$\cP_{max\;distance} = \{c | c \in C, \; c \neq a\}$$
\section{Experiments}

We ran experiments on one synthetic and four open source datasets~\cite{Dua:2019} using a fixed dense neural network over all trials.

% with  categoricals were one-hot encoded and the classifier architecture was fixed as a dense network, allowing for more variation in the choice of XAI methods. Performance of the the dense network models, as measured by AUC, is shown in comparison to logistic regression (LR) model to gauge the linearity of the datasets.

\begin{table}[ht!]
\centering
    \begin{tabular}{lrrrr}
    \toprule[1pt]

    Dataset        & Samples  & Num  & Cat & OHE \\ 
    \hline
    synthetic      &   1,000  & 15   & 5   & 30\\ 
    adult          &  48,842  & 6    &  8  & 100    \\ 
    german credit  &   1,000  & 7    & 13  & 54\\
    har            &  10,299  & 561  & 0   & 0\\
    spambase       &   4,601  & 0    & 57  &  57\\
    \hline
    \end{tabular}
    \vspace{0.1cm}
\caption{\label{tab:data}Summary of datasets, detailing number of samples, numerical (Num), categorical (Cat), and one hot encoded (OHE) features. - See \cite{Dua:2019} for open source available datasets.}
\vspace{-0.6cm}
\end{table}

% \begin{table}[ht!]
% \centering
%     \begin{tabular}{cccc}
%     \toprule[1pt]

%     Dataset       & Samples   & \# Cat./Num.  & \# Enc. Cat./Num.\\ %\midrule
%     \hline
%     synthetic       &   1,000   & 5/15 & 30/15\\ 
%     \href{https://archive.ics.uci.edu/ml/datasets/adult}{adult}           &   48,842   & 8/6 & 100/6    \\ 
%     \href{https://archive.ics.uci.edu/ml/datasets/statlog+(german+credit+data)}{german credit}   &   1,000    & 13/7    & 54/7\\
%     \href{https://archive.ics.uci.edu/ml/datasets/human+activity+recognition+using+smartphones}{human activity (har)}  &   10,299   & 0/561 & 0/561\\
%     \href{https://archive.ics.uci.edu/ml/datasets/spambase}{spambase}        &   4,601    & 0/57  & 0/57\\
%     \hline
%     \end{tabular}
%     \vspace{0.1cm}
% \caption{\label{tab:data}Summary of each dataset in the study and comparison of model performance on original datasets.}
% \vspace{-0.6cm}
% \end{table}

% \begin{table}[ht!]
% \centering
%     \begin{tabularx}{\linewidth}{crcYc}
%     \toprule[1pt]

%     Dataset       & Samples   & cat/num & LR AUC  &  NN AUC   \\ %\midrule
%     \hline
%     Synthetic       &   1,000   & 5/15    & 0.95   &   0.94  \\ 
%     \href{https://archive.ics.uci.edu/ml/datasets/adult}{Adult}           &   48,842   & 8/6     & 0.91   &   0.91 \\ 
%     \href{https://archive.ics.uci.edu/ml/datasets/statlog+(german+credit+data)}{German Credit}   &   1,000    & 13/7    & 0.81  &   0.75 \\
%     \href{https://archive.ics.uci.edu/ml/datasets/human+activity+recognition+using+smartphones}{Human Activity}  &   10,299   & 0/561   & 0.99  &   0.99 \\
%     \href{https://archive.ics.uci.edu/ml/datasets/spambase}{Spambase}        &   4,601    & 0/57    & 0.96  &   0.98 \\
%     \hline
%     \end{tabularx}
% \caption{\label{tab:data}Summary of each dataset in the study and comparison of model performance on original datasets.}
% \vspace{-0.6cm}
% \end{table}

Across the five datasets, three explanation methods, five baselines, and three perturbations for both local and global experiments with three trials, we conducted a total of 1350 experiments. We use a synthetic dataset, discussed below, as a benchmark to assess the sensitivity of an ablation study over a fixed set of hyperparameters. The following subsections summarize those comparisons.

\subsection{Synthetic dataset}

We first examine ablation results on the synthetic dataset. It contains fifteen continuous normally distributed variables and five categorical features with six uniformly sampled categories. We sample coefficients from $\cN(0,1)$ for each continuous variable and each of the 30 categorical levels. We sample labels according to
 \begin{equation*}
     y_i \sim Bernoulli\left(\frac{1}{1+\exp(-c \cdot x_i)}\right).
 \end{equation*}
 An additional four uninformative random features are added as discussed in subsection~\ref{sanity_checks}.
 
% We sample labels according to $$y_i \sim Bernoulli\left(\frac{1}{1+\exp(-c \cdot x_i)}\right)$$. An additional four uninformative random features are added as discussed in subsection~\ref{sanity_checks}.

\begin{figure}
  \includegraphics[width=.9\columnwidth]{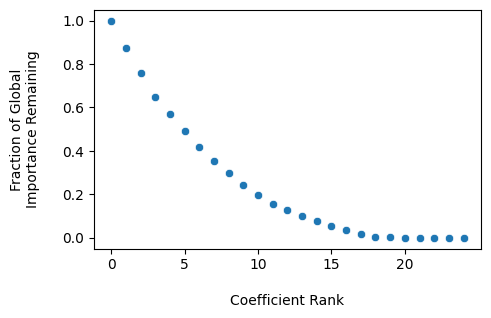}
  \caption{We train a linear model on the synthetic dataset and extract the coefficients as our ground truth global explanations. We sequentially subtract coefficients in decreasing order from the total sum to mimic the theoretical degradation of information from the model. Results of global ablation studies should mimic this decay.}
  \label{fig:synthetic_coeffs}
\end{figure}

By definition, global feature importance directly correlates to the coefficients in our data generation. Therefore, we can visualize a theoretical ablation curve based on global explanations in Figure~\ref{fig:synthetic_coeffs}. Ideally, performance should decay proportionally to the importance of the ablated features. 

\begin{figure}

  \includegraphics[width=.97\columnwidth]{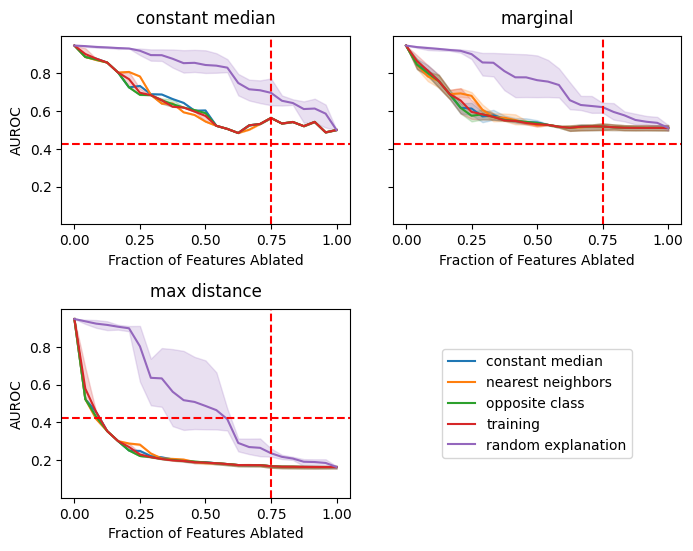}
  \caption{An example ablation study for the synthetic dataset varying both baselines and perturbations using a linear classifier, Deep SHAP explanations for global feature importance.}
  \label{fig:synthetic_linear_global}
\end{figure}

\subsection{Effect of perturbations}

To explore the impact that the choice of perturbation has on ablation, we focus on one explanation method (Deep SHAP), a variety of baselines, and vary the perturbations on the synthetic dataset, as seen in Figure~\ref{fig:synthetic_linear_global}.

It is evident that the max distance perturbation has the most significant impact on the ablation curve. It causes an extremely steep drop in performance during the ablation of the first 15\% of features, rapidly descending below the horizontal guardrail of minimum model performance. Conversely, for the other perturbations, the ablation curves remain above the guardrail and exhibit decay similar to Figure~\ref{fig:synthetic_coeffs}. It is clear that max distance is pushing the model into a decision space that deviates from those established during training.

\begin{figure}[h]
  \includegraphics[width=\columnwidth,clip]{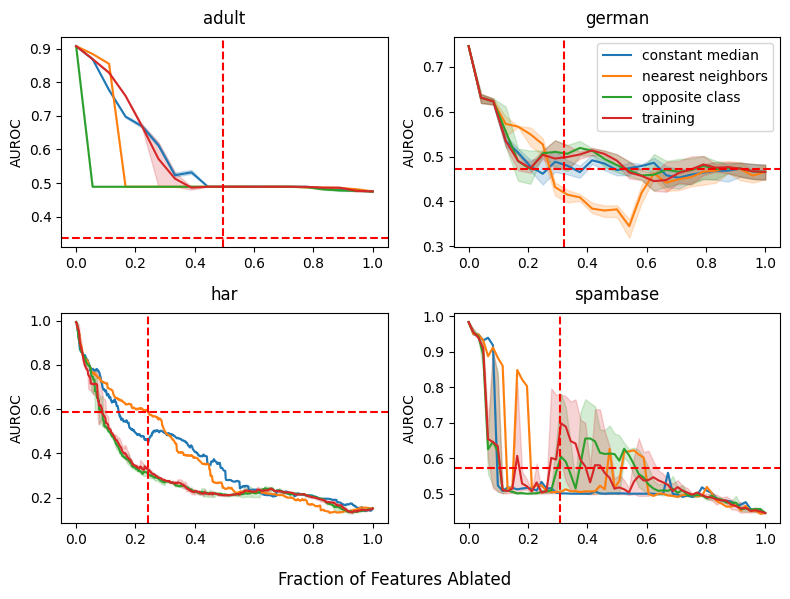}
%   \caption{The max distance perturbation results in a range of ablation curves that violate our guardrails or display unusual behavior. We perform ablation in each of these using Deep SHAP and global feature attributions.}
  \caption{The effect of the max distance perturbation across datasets and baselines, using Deep SHAP explanations for global feature importance. This extreme perturbation results in ablation curves that violate our guardrail for model capability, over-emphasize feature importance, or display wildly non-monotonic behavior.}
  \label{fig:baselines_max_dist}
\end{figure}

\begin{figure}[h]
  \includegraphics[width=\columnwidth, clip]{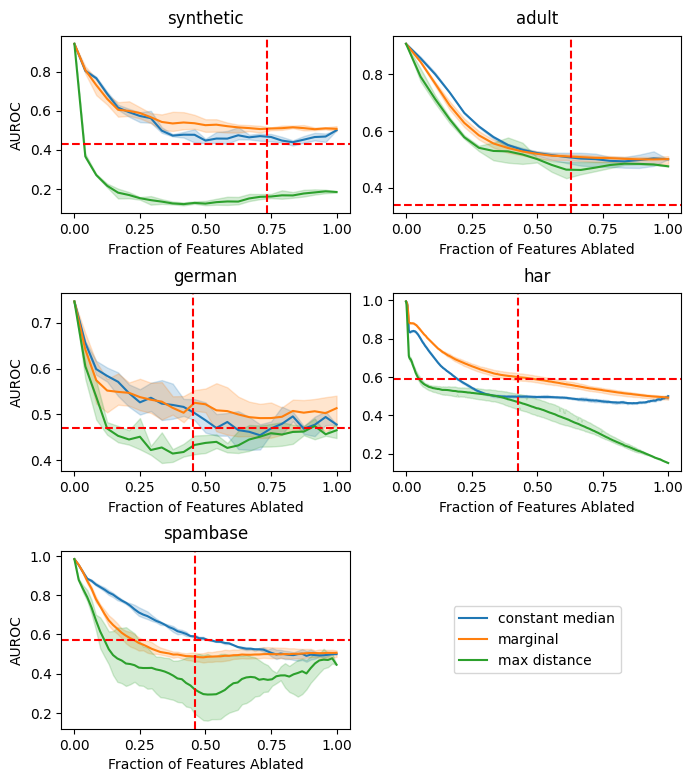}
  \caption{Comparison of ablation curves varying perturbations across the empirical datasets, using Deep SHAP, training baseline, for global importances.}
  \label{fig:perturbation_comp}
\end{figure}

This behavior is not unique to the synthetic dataset and is replicated across our empirical experiments on real datasets and explanation methods. We see steep decreases and odd behavior from har and spambase when using the max distance perturbation in Figures~\ref{fig:baselines_max_dist} and~\ref{fig:perturbation_comp}.

Interestingly, the guardrail for lower limit of model capability is sometimes, but not always, crossed by ablation experiments using the max distance perturbation (e.g., adult dataset in Figure~\ref{fig:baselines_max_dist}).
This is most likely due to a high percentage of categorical features. As discussed in Section~\ref{section:cat}, categorical features are treated more closely as a marginal sample in the max distance perturbation since categorical features already live on the hypercube. 

To summarize experiments in one metric, we focus on the infeasible area, bounded below by the ablation curve, from above by the horizontal guardrail, and on the right by the vertical guardrail. Any ablation curve that dips below the lowest capability model very quickly will incur a large area above the curve in quadrant III (see Figure~\ref{fig:sanity_quadrants}). As seen in Table~\ref{tab:area_above}, across all experiments, we find that max distance perturbation results in triple the area of any other perturbation. 

\begin{table}[h]
\centering
\begin{tabular}{ccc}
    \toprule[1pt]
    \multirow{2}{*}{\bfseries Perturbation} & 
    \multicolumn{2}{c}{\bfseries Area Measure}\\ \cline{2-3}
    & local & global \\ \hline
    %------
    constant median & 0.007 & 0.003 \\
    marginal & 0.008 & 0.004\\
    max distance & 0.024 & 0.012\\
    \bottomrule[1pt]
\end{tabular}
\vspace{0.1cm}
\caption{\label{tab:area_above} An area measure for assessing perturbations. We measure the area above ablation curves that is under the horizontal guardrail and left of the vertical guardrail in quadrant III. Areas are averaged by perturbation over all datasets, baselines, and explanation methods}
\vspace{-0.6cm}
\end{table}

\subsection{Effect of baselines}

To estimate the effect induced by a baseline, we measure the area between the random explanation curve and the ablation curve. Any time an ablation curve remains under the random explanation it contributes positively to the area, and if it crosses above, it decreases the area. The random explanation curve is obtained by ablating features in a random order. Therefore, a low area measurement indicates that the corresponding explanations are of low quality, as they perform similarly to a uniformly random assignment of importance. Over all experiments, the nearest neighbor baseline performs the worst for both local and global explanations, followed by the constant median baseline as seen in Table~\ref{tab:area_between}.

\begin{table}[h]
\centering
\begin{tabular}{ccc}
    \toprule[1pt]
    \multirow{2}{*}{\bfseries Baseline} & 
    \multicolumn{2}{c}{\bfseries Area Measure}\\ \cline{2-3}
    & local & global \\ \hline
    %------
    nearest neighbors & 0.037 & 0.087 \\ 
    constant median  & 0.081  & 0.109 \\ 
    training   & 0.195 & 0.111\\
    opposite class &  0.229 & 0.112\\
    \bottomrule[1pt]
\end{tabular}
\vspace{0.1cm}
\caption{\label{tab:area_between} Average area between random explanation and baseline curves over all datasets, perturbations, and explanation methods}
\vspace{-0.6cm}
\end{table}

\begin{figure}[h!]
% placeholder for synthetic global vs local - which baseline to use is TBD
  \includegraphics[width=\columnwidth]{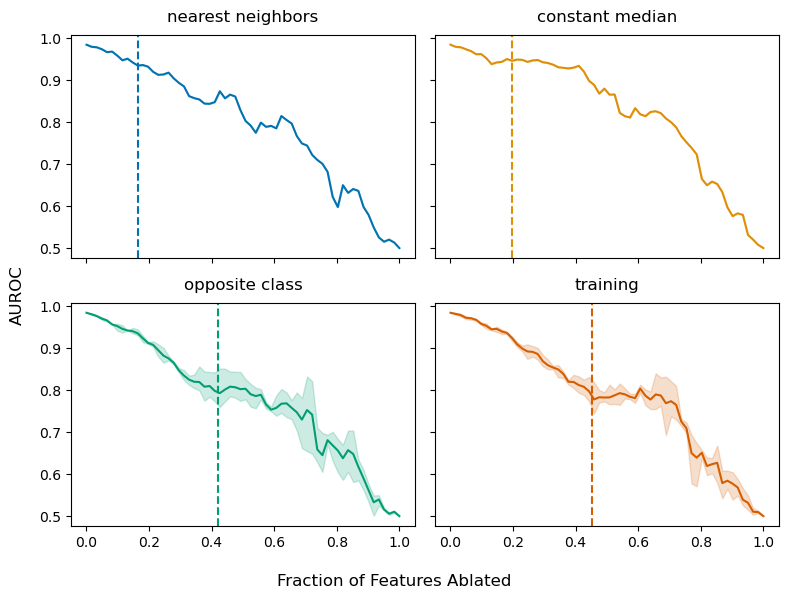}
%   \caption{Spambase dataset global guardrail comparison using Deep SHAP and constant median perturbation.}
  \caption{Comparison of global importances from different baselines using the spambase dataset, Deep SHAP explanations, and constant median perturbation. More appropriate baselines, e.g. opposite class and training, correspond with random guardrails that appear further to the right indicating more faithful explanations for non-random features.}

  \label{fig:guardrail_comp_spambase}
\end{figure}

Moreover, we hypothesize that "better" baselines should push the vertical guard to the right, indicating that the baseline helps to correctly attribute low importance to random features. This can especially be seen in Figure~\ref{fig:guardrail_comp_spambase} where less informative baselines such as nearest neighbors and constant median have guardrails closer to the left portion of the curves, while the guardrails of opposite class and training, are further to the right.

% TODO - TO BE ADDRESSED
% \todo[inline,size=\small]{
% [CAN WE SHOW THAT THESE BASELINES ARE BETTER BECAUSE THE VERTICAL GUARDRAIL IS FURTHER TO THE RIGHT? compare same dataset, same method, different baselines - the less informative baseline has the vertical guardrail to the left...]
% 
% UPDATE: Figure~\ref{fig:guardrail_comp_spambase}
% }

% [CAN WE SHOW THAT THESE BASELINES ARE BETTER BECAUSE THE VERTICAL GUARDRAIL IS FURTHER TO THE RIGHT? compare same dataset, same method, different baselines - the less informative baseline has the vertical guardrail to the left...]

\begin{figure}
  \includegraphics[width=\columnwidth]{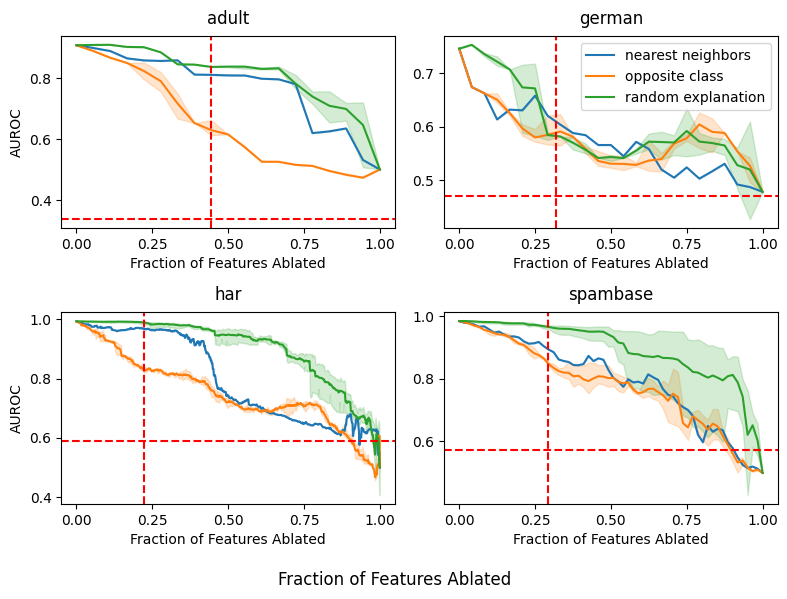}
  \caption{Comparison of the worst baseline (nearest neighbors) and best baseline (opposite class) according to our area metric, using Deep SHAP explanations and constant median perturbation for global feature importance. The opposite class curve is consistently further from the random explanation compared to nearest neighbors suggesting more informative explanations generated with opposite class baselines. See Table~\ref{tab:area_between} for details.}
  \label{fig:worst_best_baseline}
\end{figure}

\subsection{Effect of local vs. global explanations}

We conducted experiments comparing typical ablation curves and guardrails based on individual observations (local) vs across all observations (global), using various explanation methods, perturbations, and baselines. 

% Experiments for both types of explanations were run for various explanations, perturbations, and baselines. 

\begin{figure}
% placeholder for synthetic global vs local - which baseline to use is TBD
  \includegraphics[width=\columnwidth]{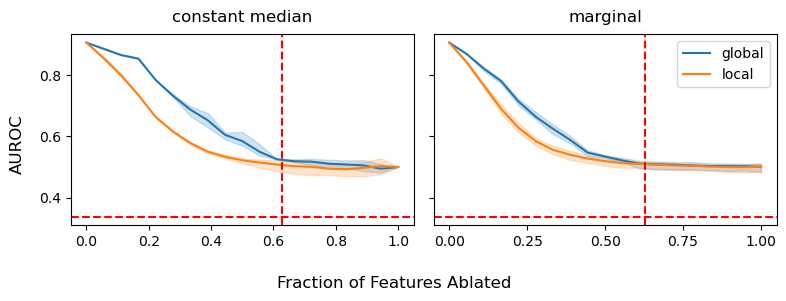}
  \caption{Comparisons of ablation curves derived from global and local explanations using the adult dataset, Deep SHAP explanations, and training baseline.}
  \label{fig:adult_global_v_local}
\end{figure}

We initially hypothesized that averaging local explanations into global explanations would produce two effects: 1) ablation curves derived from global explanations would smooth out those produced by local explanations, and 2) ablation curves produced by local explanations would initially have steeper slopes and shift curves produced by global explanations to the left. Both of these effects can be seen in Figure~\ref{fig:adult_global_v_local} with the Adult dataset. It is also worth noting that the vertical guardrails for each type of explanation are close in agreement.

% We will probably need the synthetic example of global vs local. For now, I grabbed the Adult/Deep SHAP image.

\subsection{Effect of explanation methods}
\begin{figure}[h]
  \includegraphics[width=\columnwidth]{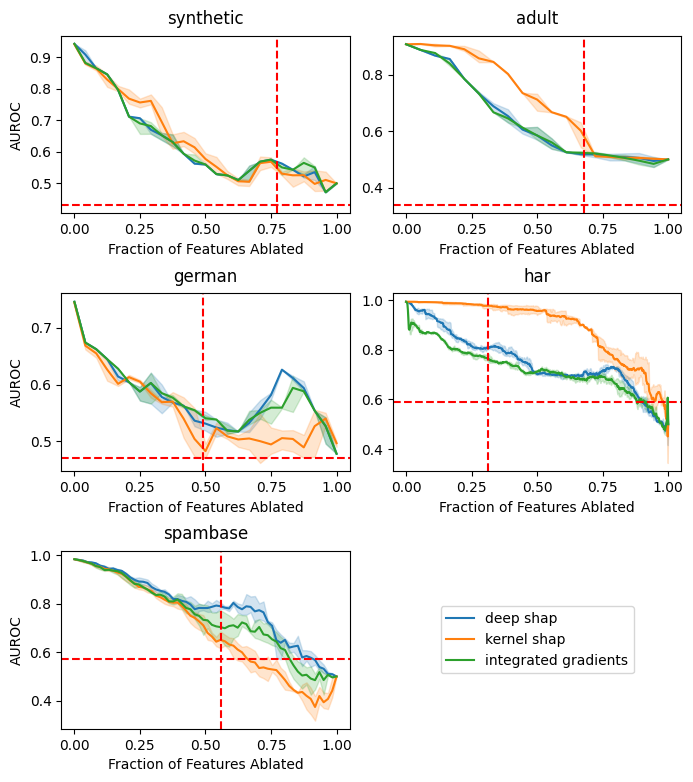}
  \caption{Comparison of ablation curves on three XAI methods across datasets indicate agreement among gradient based methods. (perturbation: constant median, baseline: training)}
  \label{fig:explanation_comparison}
\end{figure}

We explore the impact of explanation methods in ablation studies, comparing multiple explanation methods while focusing on the constant median perturbation method and training baseline, as seen in Figure~\ref{fig:explanation_comparison}. Curves produced by most explanation methods seem to agree across most datasets, with KernelSHAP as an outlier in the Adult and HAR datasets. We hypothesize that these deviations in the Adult dataset in Figure~\ref{fig:explanation_comparison} are a result of the differences in SHAP value estimation between gradient-free KernelSHAP and gradient-dependent Deep SHAP and Integrated Gradients. Since our models under test are neural networks, it is appropriate for Deep SHAP to produce higher quality explanations \cite{lundberg2017unified}. The greatest deviation for KernelSHAP occurs in HAR, see Figure~\ref{fig:explanation_comparison}. We hypothesize, and identify as an item for future work, that our baseline sample size used for KernelSHAP may be insufficient for HAR, as it has the highest feature dimensionality of all the datasets.

Differences in general trends between datasets containing categorical features (Synthetic, Adult, German), and those containing only continuous features (HAR, Spambase) could speak more to the perturbation and baseline choices rather than explanation methods. Note that selection and impact of the explanation method on the ablation curves is significantly influenced based on the choice of perturbation and baseline.

% \subsection{Categorical Treatment}
% \begin{figure*}
%   \includegraphics[width=\textwidth]{figures/KDD Specific Plots/german_agg.png}
%   \caption{Aggregating categorical features in the German Credit dataset with a constant median perturbation.}
%   \label{fig:german_agg_comparison}
% \end{figure*}

% \subsection{DOWN AND THEN BACK UP DISCUSSION - and other oddities}
% Variance in adult max distance caused by categorical randomness while the rest is a deterministic baseline. Is this a good place to bring in plots of error - as an attempt to support out of sample to in sample?

\section{Conclusions}

% What did we learn that we want the reader to take home with them. (these are our contributions)

Ablation studies can be an effective XAI validation technique with appropriate selection of perturbation method and treatment of categorical data. Our ablation analysis, augmented with guardrails as sanity checks, identified that steeper performance declines from ablation do not necessarily confirm explanation quality. The decline in model capability can be driven by perturbations that drive model inputs out of sample. Additionally, perturbations can exert an outsized influence on the ablation study, which masks the true performance of the explanations. Therefore, there is a need to carefully choose the perturbation method to avoid confounding assessments and to use other measures to ensure quality ablation studies. The three proposed guardrails are a step in that direction. They are sanity checks that help limit inappropriate conclusions drawn from an ablation analysis (e.g., the \textit{max distance} perturbation rapidly dropping model performance). 

Furthermore, data types of features must also be carefully considered. When ranking and perturbing, one-hot-encoded features should be aggregated back to their original multi-category form. Perturbations used on categorical features should maintain the data type; using any representation that falls outside of a feature's categories (e.g., mean of one hot encoding) should be avoided.
%Treatment as a single feature is especially important with regards to uninformative categorical features. 

The aggregation to the multi-categorical representation improves the utility of the random feature guardrail. The sparsity induced by one-hot-encoding creates high dimensional local attributions, with the vast majority having little importance. This results in the vertical guardrail moving further to the left, compressing the relevant region for the ablation study, and increasing the difficulty in interpreting the results as shown in Figure~\ref{fig:german_agg_comparison}.

Future work in this area should include
techniques to bound the out of distribution problem for perturbations, an expansion to include more model types, baselines, and XAI methods, and a more rigorous measurement of the lower limit of feature importance, as shown in our vertical guardrail.

% \begin{itemize}
    % \item Extension to non-differentiable models
    % \item Experimentation with actual removal (a tabular replication of ROAR with recalculation of XAI local attributions each time a model is rebuilt - or confirming that the ablated feature would have minimal importance in the newly rebuilt model)
    % \item conditional sampling to avoid the "out of distribution problem" --- 
    % \item surrogate models  (needs explanation)
    % \item more rigorous random feature treatment
    % \item Analysis of efficiency of ablation methods (as well as being in sample?) -- do we still want to do this?
    % \item Propose alternative ways perceiving ablation studies that promote calibrated explanations -- do we still want to do this?
% \end{itemize}

% REMOVE THIS WHEN FINISHED WITH PAPER
% \include{06_TODO}

\bibliographystyle{ACM-Reference-Format}
\nocite{captum}
\nocite{Dua:2019}
\bibliography{ablation}

\end{document}